\pdfoutput=1
\documentclass[11pt]{article}

\usepackage[final]{acl}

\usepackage{times}
\usepackage{latexsym}
\usepackage{multirow}
\usepackage{amsmath}
\usepackage{xcolor,soul}
\usepackage{amssymb}
\usepackage[linesnumbered]{algorithm2e}
\usepackage{amsfonts}
\usepackage{booktabs}
\usepackage{bm}
\usepackage{diagbox}
\usepackage{float}
\usepackage{fixltx2e}
\usepackage{todonotes}

\usepackage[T1]{fontenc}
\usepackage[utf8]{inputenc}

\usepackage{microtype}

\usepackage{inconsolata}

\usepackage{graphicx}

\DeclareRobustCommand{\hladv}[1]{{\sethlcolor{red!50}\hl{#1}}}
\DeclareRobustCommand{\hlchange}[1]{{\sethlcolor{orange!50}\hl{#1}}}
\DeclareRobustCommand{\hlcorrect}[1]{{\sethlcolor{green!50}\hl{#1}}}
\newcommand{\zlchange}[1]{\textcolor{black}{#1}}

\title{DiffuseDef: Improved Robustness to Adversarial Attacks via Iterative Denoising}
\author{Zhenhao Li, Huichi Zhou, Marek Rei, Lucia Specia \\
  Language and Multimodal AI (LAMA) Lab, Imperial College London \\
  \texttt{\{zhenhao.li18, h.zhou24, marek.rei, l.specia\}@imperial.ac.uk}}

\begin{document}
\maketitle
\begin{abstract}
Pretrained language models have significantly advanced performance across various natural language processing tasks. However, adversarial attacks continue to pose a critical challenge to systems built using these models, as they can be exploited with carefully crafted adversarial texts. 
Inspired by the ability of diffusion models to predict and reduce noise in computer vision, we propose a novel and flexible adversarial defense method for language classification tasks, {\it DiffuseDef} \footnote{Codes available at: \url{https://github.com/Nickeilf/DiffuseDef}}, which incorporates a diffusion layer as a denoiser between the encoder and the classifier. The diffusion layer is trained on top of the existing classifier,  ensuring seamless integration with any model in a plug-and-play manner. During inference, the adversarial hidden state is first combined with sampled noise, then denoised iteratively and finally ensembled to produce a robust text representation. By integrating adversarial training, denoising, and ensembling techniques, we show that DiffuseDef improves over existing adversarial defense methods and achieves state-of-the-art performance against common black-box and white-box adversarial attacks.

%In computer vision, diffusion models are extensively used to generate images by performing reverse diffusion steps that reduce noise from a sampled noisy image. 

%Systematic experimentation and analysis demonstrate the efficacy of DiffuseDef, affirming its potential as a universal method to improve robustness when combined with almost any adversarial defense method.
\end{abstract}

\section{Introduction}
Pretrained language models (PLM) have significantly advanced the performance of various natural language processing (NLP) tasks. Despite such improvements, current NLP systems remain susceptible to adversarial attacks where carefully crafted text perturbations can lead to incorrect model outputs \citep{alzantot-etal-2018-generating,Jin_Jin_Zhou_Szolovits_2020,li-etal-2020-bert-attack}. In order to improve robustness to adversarial attacks, various defense methods have been proposed, such as adversarial training \citep{Zhu2020FreeLB,si-etal-2021-better,zhou-etal-2021-defense,xi-etal-2022-efficient}, text denoising \citep{nguyen-minh-luu-2022-textual,wang-etal-2023-rmlm}, ensembling \citep{zhou-etal-2021-defense,zeng-etal-2023-certified,li-etal-2023-text}, etc. However, existing defense methods either assume the test-time perturbation/attack set is similar to that used in training \citep{li-etal-2021-searching}, or are limited to specific architectures \citep{xi-etal-2022-efficient}, or at inference time require large computational costs, thereby limiting their practical applicability.
%Therefore, different methods cannot be applied together to achieve complementary improvement, thereby limiting their practical applicability.

Diffusion models are commonly used in computer vision (CV) to generate high-quality images by predicting and removing noise from a sampled noisy image. Therefore, they can be adopted to remove noise from adversarial images and thus improve robustness to attacks \citep{im2016denoisingcriterionvariationalautoencoding,nie2022DiffPure,gulsen2024pcldpointcloudlayerwise}. However, in NLP very limited research has investigated adversarial defense with diffusion models due to the discrete and contextual nature of text data. \citet{li-etal-2023-text} adopt the idea of iterative denoising and reconstruct adversarial texts from masked texts, while \citet{yuan2024roicdm} use a diffusion model as a classifier and perform reverse diffusion steps on the label vector, conditioning on the input text. Inspired by the general noise prediction and reduction capability of diffusion models, we propose {\it DiffuseDef}, a novel adversarial defense method which employs diffusion training to denoise hidden representations of adversarial texts. Unlike \citet{li-etal-2023-text} and \citet{yuan2024roicdm}, which apply diffusion on texts or labels, DiffuseDef directly removes noise from the hidden states, providing a more effective and robust text representation to defend against adversarial texts. Different from diffusion-based defenses in CV \citep{nie2022DiffPure,gulsen2024pcldpointcloudlayerwise} that apply purification to input images, DiffuseDef only performs denoising at the final hidden layer, resulting in an improvement in efficiency. This improvement is further enhanced when combined with ensembling, as it eliminates the need for a forward pass through all classifier parameters for each ensemble.

DiffuseDef combines adversarial training with diffusion training, where the diffusion layer is trained to predict randomly sampled noise at a given timestep. During inference, the diffusion layer serves as a denoiser, iteratively removing noise from adversarial hidden states to yield a robust hidden representation. Moreover, we adopt the ensembling strategy by first adding random noise to the text hidden states to create multiple variants, then denoising them via the diffusion layer. The model output is made by averaging all denoised hidden states. Since ensembling happens solely at the diffusion layer, DiffuseDef is more efficient than traditional ensembling-based methods \citep{ye-etal-2020-safer,zeng-etal-2023-certified}, which require a full forward pass through all model parameters.

Through systematic experimentation, we demonstrate that DiffuseDef outperforms strong defense methods and is able to defend against multiple types of black-box and white-box adversarial attacks, while preserving performance on clean texts. Our analysis also reveals that the ensembling diffused representation provides a stronger defense against finding vulnerable words to attack and can reduce the distance in latent space between adversarial texts and their clean text counterparts.

Our contributions can be summarized as follows:
\begin{itemize}
    \item We propose DiffuseDef, a novel and flexible adversarial defense method that can be added on top of any existing adversarial defense methods to further improve robustness to adversarial attacks.
    \item DiffuseDef outperforms existing adversarial methods and achieves state-of-the-art performance against prevalent black-box and white-box adversarial attacks.
    \item Through extensive analysis, we demonstrate the effectiveness of the ensembling diffused representation and the efficiency of DiffuseDef compared to existing ensembling-based methods.
\end{itemize}

\section{Related Work}
\subsection{Textual Adversarial Attacks}
\label{sec:adv_attack}
Textual adversarial attacks focus on constructing adversarial examples from an original text that maximize the likelihood of incorrect predictions by a neural network. These attacks require adversarial examples to be perceptually similar to the original text, which is typically achieved by introducing subtle perturbations to the original text, such as character swapping \citep{JiDeepWordBug18,ebrahimi-etal-2018-adversarial}, synonym-substitutions \citep{ren-etal-2019-generating,yoo-qi-2021-towards-improving,Liu_Xu_Zhang_Xu_Zhang_Ma_Chen_Yu_Zhang_2023}, and paraphrasing \citep{gan-ng-2019-improving,huang-chang-2021-generating}. Taking the text classification task as an example, given a classifier $\mathcal{C}(\mathbf{x})$ that maps an input sequence of words $\mathbf{x} = [w_1, w_2, ..., w_L]$ to its designated label $y$, the goal of the attack model is to construct an adversarial example $\mathbf{x^{\prime}} = \mathbf{x}+ \delta$ to fool the classifier, where $\delta$ is a subtle adversarial perturbation constrained by $||\delta|| < \omega$. The adversarial example $\mathbf{x^{\prime}}$ is considered a successful attack if it leads to an incorrect prediction $\mathcal{C}(\mathbf{x}^{\prime}) \not= y$. The attacker can iteratively generate multiple adversarial examples and query the classifier to obtain a successful attack, whereas the classifier must consistently return the correct prediction within a specified number of query attempts to be considered robust.

Common textual adversarial attack methods adopt a two-stage process to construct effective adversarial examples: \textit{word importance ranking} and \textit{word substitution}. In the first stage, words or subwords are ranked based on their influence on the model's prediction. This is measured by leveraging either gradient information \citep{liu-etal-2022-character} or changes in prediction probabilities when words are removed \citep{Jin_Jin_Zhou_Szolovits_2020} or masked \citep{ren-etal-2019-generating,li-etal-2020-bert-attack}. In the second stage, candidate words are substituted with synonyms \citep{zang-etal-2020-word}, perturbed variants \citep{JiDeepWordBug18}, or outputs from masked language models \citep{garg-ramakrishnan-2020-bae,li-etal-2020-bert-attack}. The substitution process is guided by various constraints to ensure the adversarial example remains natural and semantically equivalent to the original text. Common constraints include thresholding the similarity between the replacement word embedding and the substituted word embedding, or ensuring the semantic similarity \citep{zhou-etal-2024-evaluating-validity} between sentence vectors modeled from Universal Sentence Encoder \citep{cer-etal-2018-universal}. Despite these constraints, current textual adversarial attacks still pose significant challenges to NLP models \citep{liu-etal-2022-character,xu-etal-2021-grey,yuan-etal-2023-bridge}, highlighting the necessity for defense methods for better adversarial robustness.

\subsection{Adversarial Defense Methods}
To mitigate the performance degradation caused by adversarial attacks, various adversarial defense methods have been developed. They can be grouped into three categories: \textit{training-based}, \textit{ensembling-based}, and \textit{denoising-based} methods. Adversarial training improves the robustness of the model to adversarial examples through strategies like data augmentation \citep{si-etal-2021-better,bao-etal-2021-defending} and adversarial regularization \citep{madry2018towards,Zhu2020FreeLB,wang2021infobert,xi-etal-2022-efficient,gao-etal-2023-dsrm,formento-etal-2024-semrode}. However, adversarial training methods are limited as they assume similar train-test adversarial examples, and thus tend to overfit to specific types of adversarial attacks. Ensembling-based methods generate multiple variants of the input text at inference time and ensemble model predictions over all the variants \citep{ye-etal-2020-safer,zhou-etal-2021-defense,zeng-etal-2023-certified,li-etal-2023-text}, but they can be inefficient given that model predictions are needed on every ensemble, increasing the inference time with the number of ensembles. More recently, denoising-based methods have been proposed to improve adversarial robustness by mapping the vector representation of the adversarial text to another point in the latent space that is close to the clean text \citep{nguyen-minh-luu-2022-textual,wang-etal-2023-rmlm,moon-etal-2023-randomized,yuan2024roicdm,ji-etal-2024-advancing}. The denoised representation makes it more difficult to find vulnerable words to attack, thus improving adversarial robustness \citep{wang-etal-2023-rmlm}. Nevertheless, denoising might lead to very different representations of clean text and adversarial text, therefore changing the semantic meanings.

The proposed DiffuseDef builds on these three approaches and can use any adversarially trained classifier as the base, applying denoising via a diffusion layer, and ensembling the diffused representations with a small number of ensembles. Using a diffusion layer as a denoiser addresses the overfitting problem from adversarial training and mitigates the efficiency problem by performing ensembling only at the diffusion layer. By averaging denoised hidden states across all ensembles, DiffuseDef also addresses the issue stemming from denoising, maintaining good performance on clean texts.
%Such a combination can mitigate the efficiency problem stemming from traditional ensembling methods and benefit from all three robustness approaches.

\section{{DiffuseDef}}
\begin{figure*}[htb]
\centering
  \includegraphics[width=\textwidth]{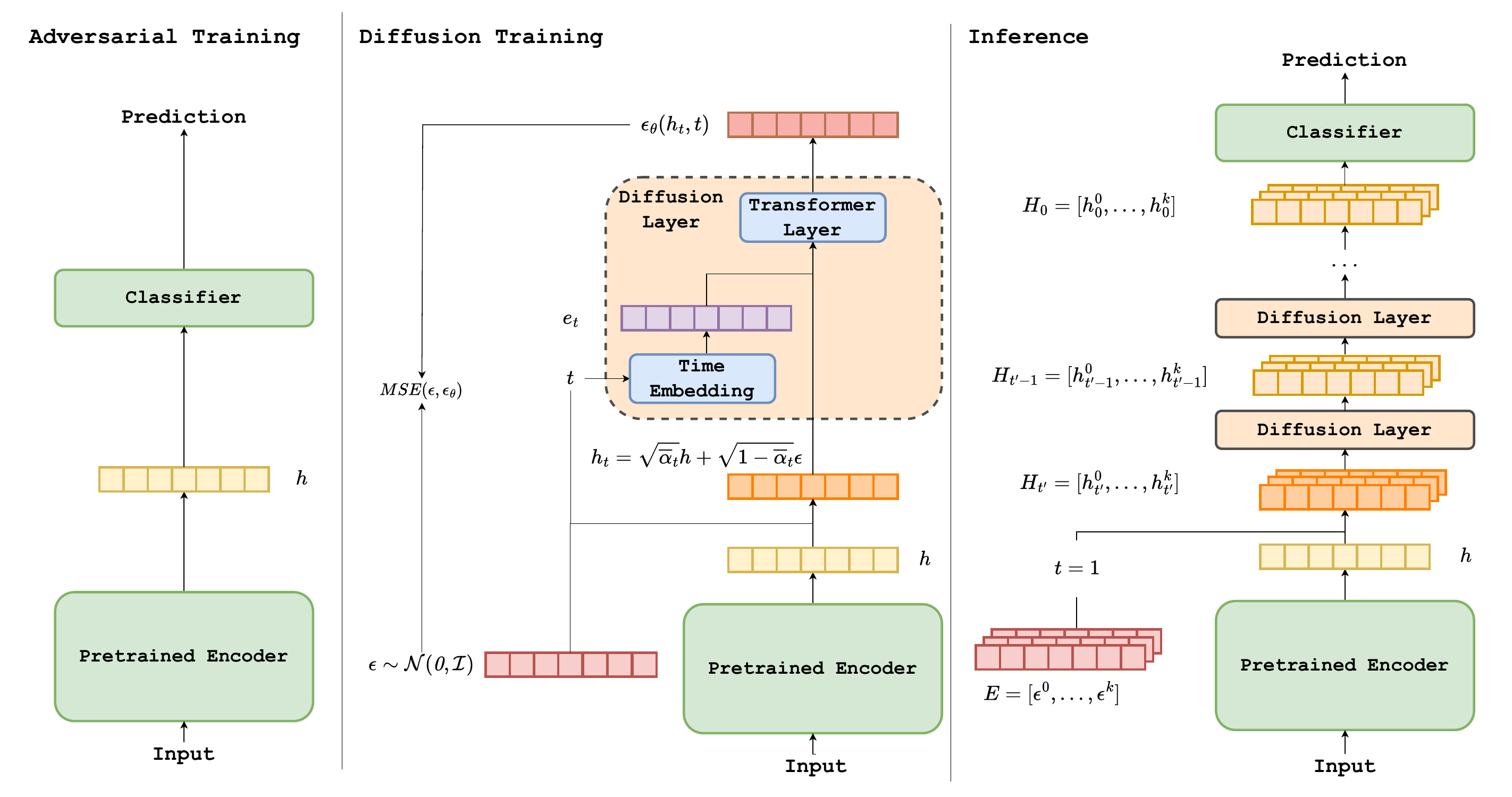}
  \caption{Training and inference of DiffuseDef model. The \textit{adversarial training} stage trains the pretrained encoder and classifier with perturbed input for adversarial robustness. The \textit{diffusion training} trains the diffusion layer to predict injected noise at a given timestep $t$. At \textit{inference} time, the text hidden state is first noised by 1 step and then denoised by $t^\prime$ steps to create the denoised hidden states, which are ensembled to make the final prediction.}
  \label{fig:model}
\end{figure*}
\subsection{Training}
The proposed diffusion defense model consists of a pretrained encoder for feature extraction, a transformer-based diffusion layer for noise prediction and reduction, and a classifier layer for output generation. The training process is split into two stages: adversarial training and diffusion training (Figure~\ref{fig:model}). The \textbf{adversarial training} stage employs a neural network-based adversarial training method like FreeLB++ \citep{li-etal-2021-searching} or RSMI \citep{moon-etal-2023-randomized}, which optimizes the encoder and classifier for robustness by perturbing the latent representation of the text input.

In the \textbf{diffusion training} stage, only the diffusion layer is trained to predict random noise added to the clean text hidden state at different timesteps, enabling it to denoise the adversarial hidden state at inference time. The pretrained encoder, however, is frozen during this stage. Since the pretrained encoder is only used for feature extraction, the diffusion training method is compatible with any neural network-based adversarial training method.

Given an input sequence of tokens $\mathbf{x} \in \mathbb{R}^L$, the pretrained encoder extracts the hidden state $h \in \mathbb{R}^{L \times D}$. A random Gaussian noise $\epsilon$ is sampled to perturb the hidden state $h$. \citet{pmlr-v37-sohl-dickstein15} define the forward diffusion process as a Markov Chain where at each timestep a Gaussian noise is sampled and added to the previous latent feature: $h_t = \sqrt{1-\beta_t} h_{t-1} + \sqrt\beta \epsilon$, where $\epsilon \in \mathcal{N}(0, \mathcal{I})$, $h_t$ is the noisy hidden state at step $t$ and $\beta$ is a pre-calculated variance schedule changing with $t$. As shown by \citet{NEURIPS2020_4c5bcfec}, this equation can be reformulated to calculate $h_t$ directly from $h$ by defining $\alpha_t = 1 - \beta_t$ and $\bar\alpha = \prod^t_{i=1}\alpha_i$, thus

\begin{equation}
    h_t = \sqrt{\bar\alpha_t}h + \sqrt{1-\bar\alpha_t}\epsilon
    \label{eq:add_noise}
\end{equation}

At each training step, a random forward diffusion timestep $t$ is sampled from a uniform distribution. Therefore, the noisy hidden state $h_t$ is created from $h$, $t$, and $\epsilon$. The diffusion layer $\theta$ consists of a time embedding and a transformer layer. The time embedding receives the diffusion timestep $t$ as input and produces an embedding $e_t$, which is added to $h_t$ as input for the transformer layer. Finally, the transformer layer outputs the predicted noise $\epsilon_\theta(h_t, t)$, and mean square error (MSE) is used to compute the loss between the predicted noise $\epsilon_\theta(h_t, t)$ and the actual sampled noise $\epsilon$.
\begin{equation}
    L = \mathbb{E}_{t,h,\epsilon}\left[\left\Vert\epsilon - \epsilon_{\theta}(\sqrt{\bar\alpha_t}h + \sqrt{1-\bar\alpha_t}\epsilon)\right\Vert^2\right]
\end{equation}

\subsection{Inference}
Leveraging the diffusion layer's ability to predict noise at a given timestep $t$, we utilize it as a denoiser during inference by iteratively performing the reverse diffusion steps, which sample from $p_\theta(h_{t-1}|h_t) = \mathcal{N}(h_{t-1}; \mu_\theta(h_t, t), \Sigma_\theta(h_t, t))$ to produce the denoised hidden state
\begin{align}
   \mu_\theta(h_t, t) &= \frac{1}{\sqrt{\alpha_t}}\left(h_t - \frac{1 - \alpha_t}{\sqrt{1-\bar\alpha_t}} \epsilon_t\right) \\
   \Sigma_\theta(h_t, t) &= \sigma^2_t\mathcal{I}
\end{align}
where $\epsilon_t$ is the predicted noise from diffusion layer and $\sigma^2_t = \beta_t$. The denoised hidden state can thus be computed with
\begin{equation}
    h_{t-1} = \frac{1}{\sqrt{\alpha_t}}\left(h_t - \frac{1 - \alpha_t}{\sqrt{1-\bar\alpha_t}} \epsilon_t\right) + \sigma_tz
\end{equation}
where $z \in \mathcal{N}(0, \mathcal{I})$.

Inference in DiffuseDef combines a one-step \textbf{noising}, a multi-step \textbf{denoising}, and an \textbf{ensembling} step. After the pretrained encoder extracts its hidden state $h$, a set of $k$ Gaussian noise vectors $E=[\epsilon^0, \epsilon^1, ..., \epsilon^k]$ are sampled to perform a single forward diffusion step. These noise vectors $E$ are then added to the hidden state $h$ following equation \ref{eq:add_noise}, resulting in a set of noisy hidden states $H_{t^\prime}=[h^0_{t^\prime}, h^1_{t^\prime}, ..., h^k_{t^\prime}]$, where ${t^\prime}$ denotes the number of denoising steps. The noisy hidden states $H_{t^\prime}$ are subsequently denoised through $t^\prime$ reverse diffusion steps, where noise is predicted by the diffusion layer and subtracted from the previous noisy hidden states. Unlike \citet{NEURIPS2020_4c5bcfec}, where the reverse diffusion step starts with pure noise sampled from a standard normal distribution, we assume the noisy hidden state $H_{t^\prime}$ is already an intermediate state in the reverse diffusion steps. This allows us to use a smaller number of $t^\prime$ than the training timestep $t$ to prevent the denoised hidden states from diverging substantially from the initial hidden state $h$. This sequence of denoising steps creates the final denoised hidden states $H_0=[h^0_0, h^1_0, ..., h^k_0]$, which are averaged and used by the classifier to output the final predicted label. This process is summarized in Algorithm \ref{alg:inference}.

\RestyleAlgo{ruled}
\begin{algorithm}[hbt!]
\caption{Inference of DiffuseDef}\label{alg:inference}
\KwData{Input text $\mathbf{x}$}
\KwResult{Predicted label $y^\prime$}
$h \gets Enc(\mathbf{x})$\;
Sample $E=[\epsilon^0, \epsilon^1, ..., \epsilon^k]$, $\epsilon \sim \mathcal{N}(0, \mathcal{I})$\;
$H_{t^\prime} \gets \sqrt{\bar\alpha_1}h + \sqrt{1-\bar\alpha_1}E$\;
\For{$i \gets 0$ \KwTo $t^\prime-1$}{
    $E_{t^\prime-i} \gets \epsilon_\theta(H_{t^\prime-i}, {t^\prime-i})$\;
    $H_{t^\prime-i-1} \gets \frac{1}{\sqrt{\alpha_{t^\prime-i}}}\left(H_{t^\prime-i} - \frac{1 - \alpha_{t^\prime-i}}{\sqrt{1-\bar\alpha_{t^\prime-i}}} E_{t^\prime-i}\right) + \sigma_{t^\prime-i}z$\;
}
$y^\prime \gets CLS\left(avg(H_0)\right)$\;
\end{algorithm}

\section{Experiments}
\paragraph{Datasets.}
We focus on two common NLP tasks in our experiments: topic classification and natural language inference (NLI). We conduct experiments on three common datasets: AG News \citep{agnews2015}, IMDB \citep{maas-etal-2011-learning}, and QNLI datasets \citep{wang-etal-2018-glue}. We randomly split AGNews, IMDB, and QNLI datasets into train, validation, and test splits. Appendix \ref{sec:data_prep} gives details on data preparation.

\paragraph{Evaluation.}
Following previous work on adversarial defense, we use three benchmarking black-box attack methods to evaluate the robustness of DiffuseDef: TextFooler (TF) \citep{Jin_Jin_Zhou_Szolovits_2020}, TextBugger (TB) \citep{DBLP:conf/ndss/LiJDLW19}, and Bert-Attack (BA) \citep{li-etal-2020-bert-attack}. %The three attack methods create adversarial attacks in different granularities: character-level perturbation (TextBugger), word substitution (TextFooler), and subword substitution (BertAttack). 
We also evaluated the robustness against two white-box attacks: T-PGD \citep{yuan-etal-2023-bridge} and SemAttack \citep{wang-etal-2022-semattack}.
Regarding evaluation metrics, we measure the clean accuracy (\textbf{Clean\%}) on the test set, the accuracy under attack (\textbf{AUA\%}), and the number of adversarial queries (\textbf{\#Query}) needed for a successful attack. Higher scores on the three metrics denote a better robustness performance of a defense method. The accuracy on clean data is measured across the entire test set. The accuracy under attack and number of queries, due to the lengthy attacking process, is measured on a randomly sampled subset of 1000 examples from the test set. We use the {\tt TextAttack} %\footnote{\url{https://github.com/QData/TextAttack}} \citep{morris-etal-2020-textattack} 
library as the adversarial evaluation framework. To ensure a fair comparison and high-quality adversarial examples, we follow the same evaluation constraints as in \citet{li-etal-2021-searching}. 
%Detailed constraints on each dataset are shown in the Appendix \ref{sec:eval_param}. 
The evaluation metrics are averaged based on experiments run with 5 random seeds. % to avoid potential improvement stemming from one specific random seed.

\subsection{Comparison to SOTA}
We compare our proposed method with state-of-the-art adversarial defense approaches, trained using both BERT \citep{devlin-etal-2019-bert} and RoBERTa \citep{liu2019roberta} as backbones: \textbf{Fine-tune}: Fine-tuning pretrained models on downstream tasks with no defense method applied\footnote{\zlchange{``Fine-tuned'' is a baseline approach used to illustrate the effect of adversarial attacks.}}. \textbf{InfoBERT} \citep{wang2021infobert}: Applying mutual-information-based regularizer during fine-tuning of pretrained models to improve robustness. \textbf{FreeLB++} \citep{li-etal-2021-searching}: An adversarial training method improving on FreeLB\citep{Zhu2020FreeLB}, which adds adversarial perturbations to word embedding during fine-tuning. \textbf{EarlyRobust}\footnote{We only run EarlyRobust with BERT as its implementation with RoBERTa has not been released.} \citep{xi-etal-2022-efficient}: Extracting early-bird subnetworks and pruning pretrained models for efficient adversarial training. \textbf{RMLM} \citep{wang-etal-2023-rmlm}: A denoising-based model combined with adversarial text detection. \textbf{ATINTER} \citep{gupta2023don}: A fine-tuned T5 model that rewrites the input adversarial texts, and the classifier predicts labels from the rewritten texts. \textbf{RSMI} \citep{moon-etal-2023-randomized}: A two-stage training method that combines randomized smoothing and masked inference to improve adversarial robustness. 

\subsection{Implementation and Settings}
We train two DiffuseDef variants using FreeLB++ and RSMI models as base models considering their robust adversarial defense capabilities. In the diffusion layer, only one transformer encoder layer \citep{NIPS2017_3f5ee243} is used. The maximum noising timestep $t$ during training is set to 30 for AGNews and QNLI datasets, and 10 for the IMDB dataset, while at inference time, we only apply 5 denoising steps for $t^\prime$. We follow \citep{NEURIPS2020_4c5bcfec} to use a linear $\beta_t$ schedule from $\beta_1=10^{-4}$ to $\beta_t = 0.02$. The base encoder and classifier are first fine-tuned for 10 epochs, and the diffusion layer is trained for 100 epochs, with the base encoder and classifier parameters frozen for efficiency. During the diffusion training stage, the same train-dev splits are used as in the adversarial training stage, thus ensuring no data leakage. At inference time, the number of ensembles is set to 10. Appendix \ref{sec:training_param} lists the training hyperparameters.

\section{Results and Analysis}
\subsection{Adversarial Robustness}

\begin{table*}[htb]
    \centering
    \small
    \resizebox{.98\linewidth}{!}{
    \begin{tabular}{c|l|c|c c c|c c c}
    \toprule
    % Table top rows
    \multirow[m]{2}{*}{ \textbf{Dataset} } & \multirow[m]{2}{*}{ \textbf{Method} } & \multirow[m]{2}{*}{\textbf{Clean\%}} & \multicolumn{3}{c|}{\textbf{AUA\%}} & \multicolumn{3}{c}{\textbf{\#Query}} \\ 
    & & & TF & TB & BA & TF & TB & BA \\
    \midrule 
    \multirow{9}{*}{ AGNews } & Fine-Tuned &  94.4 & 10.2 & 25.4 & 27.1 & 348 & 372 & 379 \\
     & InfoBERT~\citep{wang2021infobert} & 95.0 & 35.5 & 39.1 & 42.6 & 377 & 397 & 397 \\
     & FreeLB++~\citep{li-etal-2021-searching} & 95.0 & 54.7 & 56.5 & 44.6 & 426 & 430 & 390 \\
     & EarlyRobust~\citep{xi-etal-2022-efficient} & 94.4 & 35.6 & 37.2 & 45.7 & 475 & 516 & 533 \\
     & RMLM~\citep{wang-etal-2023-rmlm} & 94.3 & 45.6 & 44.4 & 54.1 & 642 & 738 & 705 \\
     % & ADFAR~\citep{bao2021defending} & \\
     & ATINTER~\citep{gupta2023don} & 94.2 & 68.0 & 59.0 & 81.0 & 527 & 235 & 122 \\
     & RSMI~\citep{moon-etal-2023-randomized} & 94.3 & 52.6 & 56.7 & 55.4 & 680 & 737 & 687 \\
     & DiffuseDef-FreeLB++ (Ours) & 94.8 & \textbf{84.5} & \textbf{86.0} & \textbf{84.6} & 877 & 972 & 910 \\
     & DiffuseDef-RSMI (Ours) & 93.8 & 82.7 & 83.3 & 84.4 & \textbf{894} & \textbf{1029} & \textbf{930} \\ 
    % \cmidrule(r){2-10}
    %  & \multirow{6}{*}{ RoBERTa-base } & Fine-Tuned & 94.9 & 34.1 & 36.9 & 43.6 & 372 & 396 & 410 \\
    %  & & InfoBERT & 95.5 & 40.2 & 45.2 & 48.6 & 392 & 421 & 430 \\
    %  & & FreeLB++ & 95.4 & 57.5 & 62.9 & 55.9 & 444 & 467 & 447 \\
    %  & & RSMI & 93.1 & 64.2 & 66.4 & 67.4 & 774 & 861 & 808 \\
    %  & & DiffuseDef-FreeLB++ (Ours) & 95.3 & \textbf{85.6} & \textbf{87.6} & \textbf{85.3} & 880 & \textbf{976} & 906 \\
    %  & & DiffuseDef-RSMI (Ours) & 92.9 & 82.9 & 83.5 & 82.2 & \textbf{905} & 925 & \textbf{1047} \\
    \midrule
    \multirow{9}{*}{ IMDB } & Fine-Tuned & 93.3 & 7.7 & 8.3 & 10.5 & 540 & 534 & 378 \\
     & InfoBERT~\citep{wang2021infobert} & 93.9 & 29.2 & 25.4 & 30.7 & 642 & 644 & 390 \\
     & FreeLB++~\citep{li-etal-2021-searching} & 94.3 & 44.2 & 39.6 & 40.6 & 784 & 829 & 426 \\
     & EarlyRobust~\citep{xi-etal-2022-efficient} & 92.7 & 49.7 & 46.8 & 43.8 & 2267 & 2788 & 1841 \\
     & RMLM~\citep{wang-etal-2023-rmlm} & 93.6 & 48.8 & 47.9 & 47.1 & 2554 & 3258 & 2064\\
     % & ADFAR~\citep{bao2021defending} & 92.9 & 34.0 & 25.9 & 25.5 & 3339 & 4324 & 2593 \\
     & ATINTER~\citep{gupta2023don} & 94.3 & 25.0 & 15.0 & 39.0 & 1173 & 645 & 637 \\
     & RSMI~\citep{moon-etal-2023-randomized} & 90.9 & 60.0 & 54.4 & 51.1 & 2840 & 3455 & 2070 \\
     & DiffuseDef-FreeLB++ (Ours) & 94.4 & \textbf{82.1} & \textbf{83.0} & \textbf{84.0} & 3174 & 4348 & 2842 \\
     & DiffuseDef-RSMI (Ours) & 90.2 & 80.9 & 79.8 & 79.8 & \textbf{3590} & \textbf{4748} & \textbf{2901} \\ 
    % \cmidrule(r){2-10}
    %  & \multirow{6}{*}{ RoBERTa-base } & Fine-Tuned & 94.6 & 21.3 & 17.9 & 13.6 & 587 & 671 & 493 \\
    %  & & InfoBERT & 94.8 & 30.9 & 27.9 & 21.8 & 681 & 760 & 549 \\
    %  & & FreeLB++ & 95.3 & 46.0 & 42.1 & 33.9 & 829 & 974 & 637 \\
    %  & & RSMI & 92.7 &  77.9 & 74.3 & 70.6 & 3443 & 4342 & 2619 \\
    %  & & DiffuseDef-FreeLB++ (Ours) & 95.0 & \textbf{86.2} & \textbf{85.9} & \textbf{86.8} & 3573 & 4663 & 2941 \\
    %  & & DiffuseDef-RSMI (Ours) & 92.4 & 84.7 & 84.1 & 84.3 & \textbf{3673} & \textbf{4782} & \textbf{3007} \\
    \midrule
    \multirow{8}{*}{ QNLI } & Fine-Tuned & 90.8 & 21.5 & 15.5 & 13.3 & 195  & 206 & 177 \\
     & InfoBERT~\citep{wang2021infobert} & 91.2 & 27.8 & 22.8 & 18.7 & 217 & 232 & 201 \\
     & FreeLB++~\citep{li-etal-2021-searching} & 90.3 & 45.6 & 40.2 & 30.5 & 253  & 279 & 226 \\
     & EarlyRobust~\citep{xi-etal-2022-efficient} & 89.2 & 24.3 & 21.2 & 19.1 & 265 & 292 & 255 \\
     % & ADFAR~\citep{bao2021defending} & \\
     & ATINTER~\citep{gupta2023don} & 90.4 & 43.0 & 24.8 & 34.0 & 185 & 115 & 89\\
     & RSMI~\citep{moon-etal-2023-randomized} & 87.4 & 35.2 & 30.9 & 28.2 & 314 & 353 & 314 \\
     & DiffuseDef-FreeLB++ (Ours) & 90.3 & \textbf{66.7} & \textbf{65.3} & \textbf{64.4} & \textbf{485} & \textbf{587} & \textbf{543} \\
     & DiffuseDef-RSMI (Ours) & 86.4 & 55.5 & 54.8 & 57.2 & 459 & 569 & 528 \\
    % \cmidrule(r){2-10}
    %  & \multirow{6}{*}{ RoBERTa-base } & Fine-Tuned & 92.8 & 26.6 & 22.5 & 20.3 & 204 & 219 & 188 \\
    %  & & InfoBERT & 92.5 & 29.1 & 25.8 & 20.3 & 205 & 223 & 189 \\
    %  & & FreeLB++ & 92.8 & 33.9 & 27.7 & 20 & 227 & 244 & 200 \\
    %  & & RSMI & 89.3 & 38.8 & 33.2 & 30.0 & 340 & 375 & 343 \\
    %  & & DiffuseDef-FreeLB++ (Ours) & 92.7 & 64.6 & 64.3 & 61.5 & 473 & 579 & 524 \\
    %  & & DiffuseDef-RSMI (Ours) & 88.8 & 57.7 & 55.7 & 53.9 & 469 & 578 & 518 \\
    \bottomrule
    \end{tabular}
    }
    \caption{Main adversarial robustness results on classification tasks with BERT. \textit{Clean}: accuracy on clean test set. \textit{TF}: TextFooler. \textit{TB}: TextBugger. \textit{BA}: BertAttack.}
    \label{tab:main_results}
\end{table*}

In Table \ref{tab:main_results}, we compare the adversarial robustness of DiffuseDef with baselines and SOTA methods on AGNews, IMDB, and QNLI datasets trained with BERT (results for RoBERTa in Appendix \ref{sec:results_roberta}). DiffuseDef consistently outperforms all other methods on all three datasets across the three attacks, exhibiting substantial improvements in accuracy under attack. After applying diffusion training, the AUA score for both FreeLB++ and RSMI models improves significantly, with an average increase of 30\% AUA against the three attack methods. When comparing the clean accuracies to its base model (i.e., FreeLB++ and RSMI), DiffuseDef only shows a minor decline, between 0.2 and 0.7 accuracy score, which indicates that it can preserve the clean text performance while improving adversarial robustness. Moreover, models trained with DiffuseDef show a much smaller gap between clean accuracy and accuracy under attack, and such a difference can be reduced to less than 10\% AUA.

Another benefit of DiffuseDef is the increased number of adversarial queries needed to obtain a successful attack. Models applying DiffuseDef require over twice the number of queries on all three datasets compared to the other methods. This increase is even larger on the IMDB dataset due to the longer text length. For example, the DiffuseDef model requires on average over 3000 queries to achieve a successful attack, while FreeLB++ can only defend for 400 to 800 queries. The substantial increase suggests that even if the attackers manage to construct a successful adversarial attack, they need 2x to 3x more time to find the attack on DiffuseDef than other models, affirming the improved robustness from diffusion training. In addition, we observe that the number of queries for denoising-based methods (i.e., RSMI, RMLM, and DiffuseDef) is generally higher than adversarial training-based methods (i.e., InfoBERT, FreeLB++). This is because denoising-based methods transform the hidden representations of the adversarial texts into a non-deterministic representation. The introduction of randomness in hidden states results in uncertainty in model logits, thus increasing the difficulty of finding vulnerable words to attack \citep{wang-etal-2023-rmlm}. 

\subsection{Robustness against White-box Attacks}
In addition to black-box attacks, we also measure the robustness against white-box attacks, where the attackers can access the defender model's gradients and parameters, which is a more challenging adversarial attack scenario than black-box attacks. In Table \ref{tab:white-box-results}, we report AUA score under T-PGD \citep{yuan-etal-2023-bridge} and SemAttack \cite{wang-etal-2022-semattack}, the former being a variant of PGD attack \citep{madry2019deeplearningmodelsresistant} for text, while the latter adopts C\&W attack \citep{carlini2017evaluatingrobustnessneuralnetworks} on embedding layers. We observe that white-box attacks result in lower AUA for all defense methods. For example, FreeLB++ performs poorly on IMDB when attacked by SemAttack, with the AUA score being 0.03. However, applying diffusion training still significantly improves adversarial robustness to white-box attacks. Note that even RSMI shows very robust performance against T-PGD\footnote{RSMI is incompatible with SemAttack because it requires gradients of the embedding layer, but SemAttack skips the embedding layer by sending embeddings directly to the target model.}, it can still further improve from DiffuseDef.

\begin{table}[htb]
    \centering
    \small
    \resizebox{\linewidth}{!}{
    \begin{tabular}{l|c c|c c}
    \toprule
        \multirow[m]{2}{*}{ \textbf{Method} }&  \multicolumn{2}{c|}{\textbf{AGNews}} &  \multicolumn{2}{c}{\textbf{IMDB}} \\
         & T-PGD & Sem & T-PGD & Sem \\
    \midrule
        Fine-Tuned & 8.8 & 41.5 & 3.0 & 1.3 \\
        InfoBERT & 32.1 & 47.1 & 25.4 & 2.0 \\
        FreeLB++ & 19.6 & 58.2 & 15.5 & 3.4 \\
        RSMI & 79.6 & n/a & 43.3 & n/a \\
        DiffuseDef-FreeLB & 59.4 & \textbf{68.1} & \textbf{50.3} & \textbf{28.2} \\
        DiffuseDef-RSMI & \textbf{81.7} & n/a & 48.4 & n/a \\
    \bottomrule
    \end{tabular}
    }
    \caption{Accuracy under white-box attacks.}
    \label{tab:white-box-results}
\end{table}

\subsection{Ablation analysis}
To understand how each component contributes to DiffuseDef, we conduct an ablation analysis on the AGNews dataset by disabling ensembling, diffusion denoising, and adversarial training one at a time. Table \ref{tab:ablation} shows the ablation results, and we find that removing either component (e.g., ensembling and diffusion denoising) at inference time leads to a sharp decrease in accuracy under attack, which indicates that both denoising and ensembling contribute significantly to the improved robustness. Nevertheless, removing adversarial training, though also introducing a degraded AUA score, is less detrimental than the other inference time components, showing that DiffuseDef is not dependent on a specific adversarial training method.
\begin{table}[htb]
    \centering
    \small
    \resizebox{\linewidth}{!}{
        \begin{tabular}{l|c|c|c}
        \toprule
            \textbf{Method} & \textbf{TF} & \textbf{TB} & \textbf{BA}  \\
        \midrule
            DiffuseDef & 84.5 & 86.0 & 84.6 \\
            \quad w/o ensembling & 64.2 & 65.8 & 54.9 \\
            \quad w/o diffusion denoising & 57.1 & 58.4 & 47.7 \\
            \quad w/o adv training & 80.3 & 80.9 & 80.5 \\
        \bottomrule
        \end{tabular}
    }
    \caption{Ablation results for DiffuseDef's accuracy under attack on AGNews datasets.}
    \label{tab:ablation}
\end{table}

\subsection{DiffuseDef + Adversarial Augmentation}
\begin{figure}[htb]
    \centering
    \includegraphics[width=\linewidth,trim={1cm 0 1cm 1.6cm},clip]{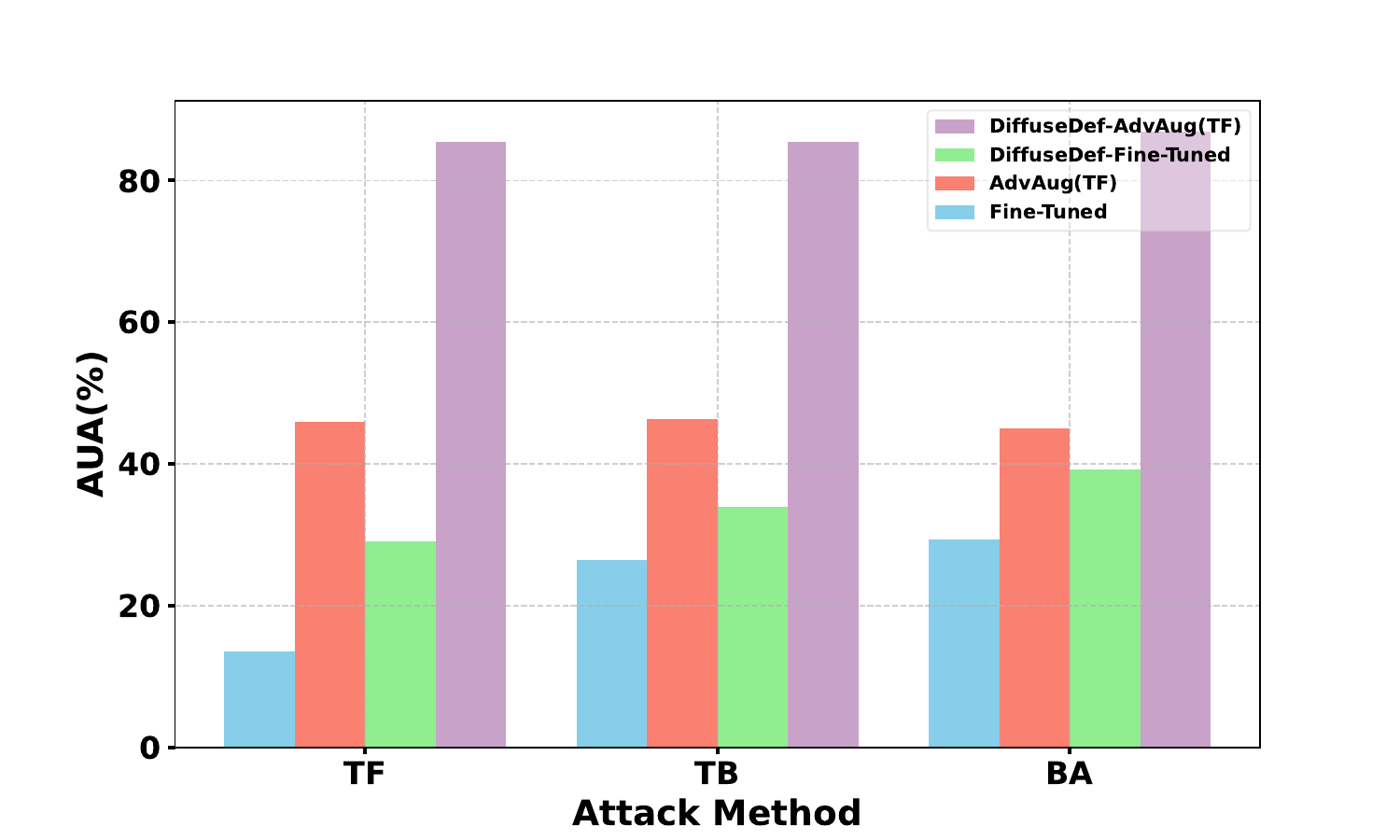}
    \caption{Robustness of DiffuseDef with textual adversarial augmentation method.}
    \label{fig:augmentation-results}
\end{figure}

In Figure~\ref{fig:augmentation-results}, we combine DiffuseDef with adversarial data augmentation and evaluate the robustness on AGNews. We first generate adversarial augmentation data using TextFooler and adversarially train the classifier with augmented training data. The augmented classifier is further performed with diffusion training, resulting in a DiffuseDef variant. We notice that adversarial training on augmented data helps improve the performance on all three attacks. After applying diffusion training, the adversarial performance significantly improves over the augmented model, which proves the flexibility of DiffuseDef to be applied with different types of adversarial training methods.
% \begin{table}[htb]
%     \centering
%     \small
%     \begin{tabular}{l|c c c}
%     \toprule
%         \textbf{Method}& \textbf{TF} & \textbf{TB} & \textbf{BA} \\
%     \midrule
%         Base & 13.6 & 26.4 & 29.3 \\
%         AdvAug(TF) & 45.9 & 46.3 & 45.0 \\
%         DiffuseDef-base & 29.1 & 34.0 & 39.3 \\
%         DiffuseDef-AdvAug(TF) & 85.4 & 85.5 & 86.9 \\
%     \bottomrule
%     \end{tabular}
%     \caption{Robustness of DiffuseDef with textual adversarial augmentation method.}
%     \label{tab:augmentation-results}
% \end{table}

\subsection{Effect of Additional Denoising Steps}
\begin{figure}[htb]
    \centering
    \includegraphics[width=\linewidth,trim={1cm 0 1cm 0},clip]{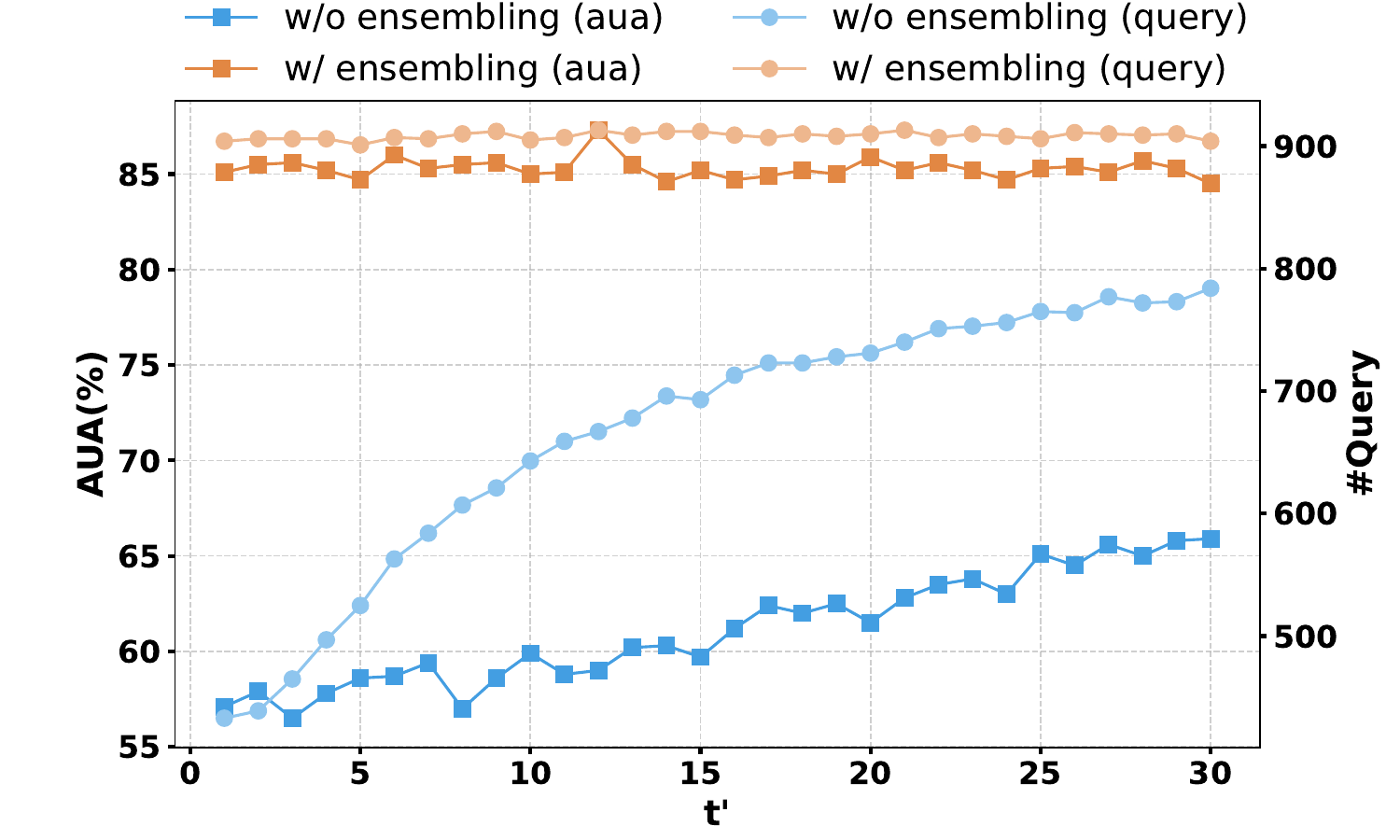}
    \caption{AUA and \#Query (TextFooler) w.r.t inference denoising step for DiffuseDef w/ and w/o ensembling.}
    \label{fig:inference_step}
\end{figure}

In Figure \ref{fig:inference_step}, we study how the inference denoising steps $t^\prime$ can affect the adversarial performance. For the DiffuseDef model without ensembling, both the AUA score and the number of queries required to attack increase as the inference denoising step is larger. As the denoising step $t^\prime$ grows from 1 to 30, the AUA score improves from 58 to 65 while the number of attack queries grows from 430 to 780. In contrast, for DiffuseDef with ensembling, the model maintains a stable but robust performance in AUA and number of queries, regardless of the increase of $t^\prime$. Considering that the ensembling introduces a notable performance increase, the DiffuseDef model is likely to be hitting an upper bound in both metrics, thus no further improvement is reached by increasing the denoising steps. However, it also shows that with ensembling, DiffuseDef can be applied with a smaller $t^\prime$ for better efficiency while maintaining a robust adversarial performance.

\subsection{Ensembling Diffused Hidden Representations}
In DiffuseDef, the text hidden state is diffused and ensembled to form a denoised hidden representation, which contributes significantly to the improved adversarial robustness. In this section, we study how the ensembling diffused hidden representation helps defend against adversarial attacks.
\begin{figure}[htb]
    \centering
    \includegraphics[width=\linewidth,trim={1cm 0 1cm 1.6cm},clip]{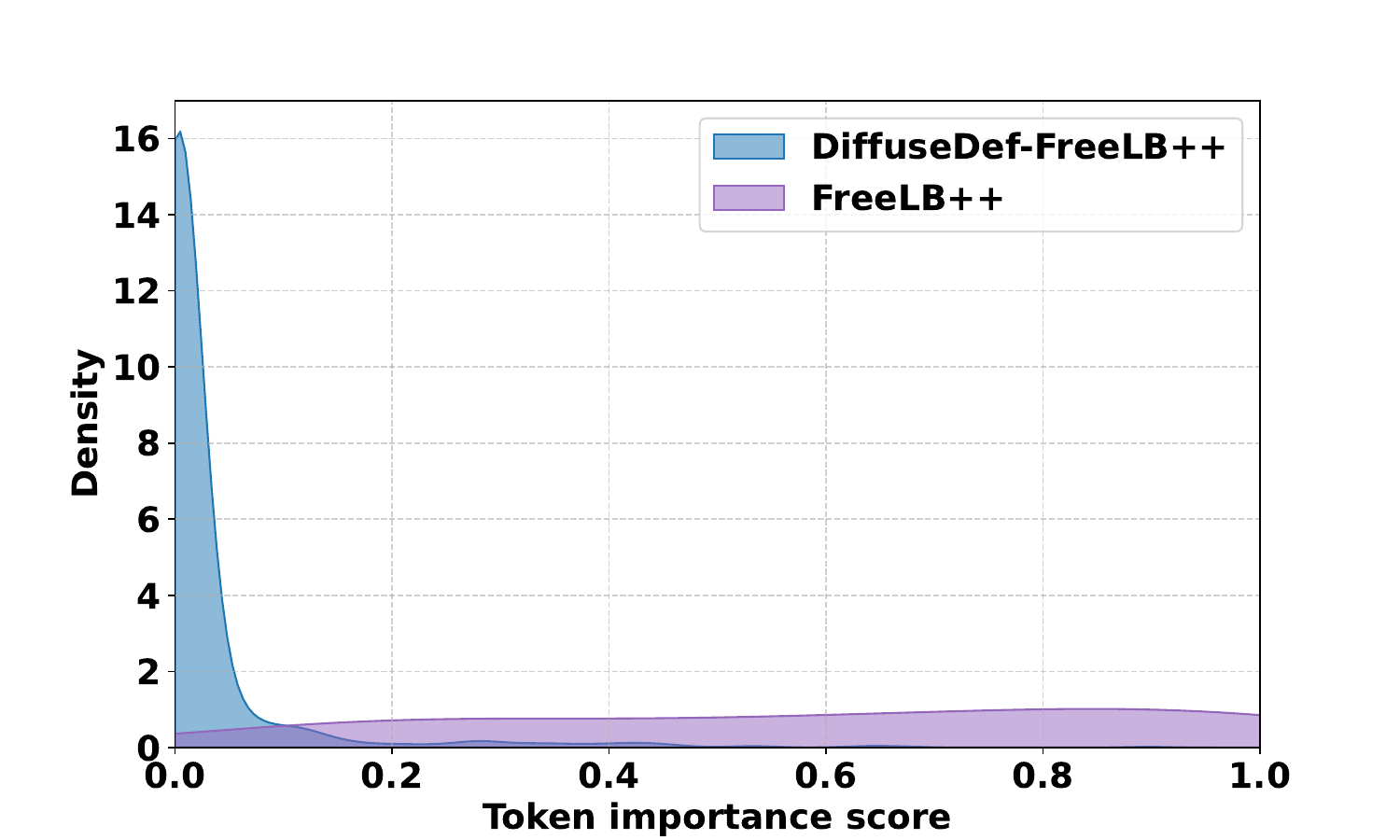}
    \caption{Distribution of max token importance score in the AGNews test set.}
    \label{fig:token_importance}
\end{figure}

As mentioned in Section \ref{sec:adv_attack}, attack methods need to first rank token importance based on its influence on prediction. Specifically, the importance score is calculated by comparing the changes in model prediction probabilities after removing each word. In Figure \ref{fig:token_importance}, we compare the density distribution of the max token importance score between FreeLB++ and its DiffuseDef counterpart. %Both FreeLB++ and DiffuseDef show a long-tail distribution with over 80 percent examples having a max token importance score below 0.1. This suggests that in most cases changing one single token will not significantly alter the prediction for both models. 
DiffuseDef shows a notably lower percentage than FreeLB++ when the max importance score is high, where the attacker can easily find the vulnerable token to construct adversarial examples. This difference shows that DiffuseDef can complicate the process of important word searching, which accounts for the increased number of queries required for a successful attack.

\begin{table}[htb]
    \centering
    \small
    \resizebox{.8\linewidth}{!}{
    \begin{tabular}{l|c|c}
    \toprule
        \textbf{Method} & \textbf{L2} & \textbf{Cosine} \\
    \midrule
        FreeLB++ & 12.53 & 0.35 \\
        DiffuseDef-FreeLB++ & 10.66 & 0.27 \\
    \midrule
        RSMI & 9.72 & 0.24  \\
        DiffuseDef-RSMI & 8.61 & 0.21 \\
    \bottomrule
    \end{tabular}
    }
    \caption{L2 and cosine distance between hidden states for clean and adversarial texts.}
    \label{tab:distance}
\end{table}
In addition, DiffuseDef mitigates the difference between clean and adversarial texts by reducing the distance between their hidden states. In Table \ref{tab:distance}, we report the L2 and cosine distance between clean and adversarial hidden states for FreeLB++ and RSMI. Both show lower L2 and cosine distance after applying DiffuseDef, indicating that ensembling diffused representation repositions the adversarial example closer to the clean example, leading to the model maintaining its predictions.

\subsection{Efficiency of DiffuseDef}

\begin{table}[htb]
    \centering
    \small
    \resizebox{\linewidth}{!}{
    \begin{tabular}{l|c|c|c}
    \toprule
        \textbf{Method} & \textbf{Params} & \textbf{FLOPS} & \textbf{Train}\\
    \midrule
        Fine-Tuned (BERT) & 110M & 46G & 1x \\
    \midrule
        EarlyRobust & 82M & 32G & 0.8x \\
        FreeLB++ & 110M & 46G & 10.5x \\
        InfoBERT & 110M & 46G & 6.5x \\
        RSMI & 110M & 92G & 1.25x \\
        RanMask ($k=10$) & 110M & 459G & 1.2x \\
        SAFER ($k=10) $& 110M & 459G & 1x\\
    \midrule
        DiffuseDef ($t^\prime=1, k=10$) & 120M & 96G & 1.1x \\
        DiffuseDef ($t^\prime=5, k=10$) & 120M & 267G & 1.1x \\
        % DiffuseDef ($t^\prime=30, k=0$) & 120M & 179G \\
    \bottomrule
    \end{tabular}
    }
    \caption{Efficiency comparison of DiffuseDef and other methods. Params: number of model parameters. FLOPS: number of floating point operations per second at inference time, calculated with batch size of 1 and sequence length of 256. Train: training time.}
    \label{tab:efficiency}
\end{table}

Given that DiffuseDef adds additional denoising and ensembling steps during inference, it inevitably increases the computation time compared to its base model. To study its efficiency, we report the number of model parameters, inference FLOPS, and training time in Table \ref{tab:efficiency}. In addition to the defense methods in Table \ref{tab:main_results}, we also compare the efficiency of DiffuseDef with two other SOTA ensembling-based defense methods, i.e., RanMask \citep{zeng-etal-2023-certified} and SAFER \cite{ye-etal-2020-safer}.

%On the number of model parameters, 
All SOTA models have the same number of parameters as the fine-tuned BERT model, except EarlyRobust, which applies attention head pruning for better efficiency. DiffuseDef requires more inference FLOPS than non-ensembling-based baselines such as FreeLB++ and EarlyRobust, due to the diffusion steps. With $t^\prime=1$ and $k=10$, the FLOPS for DiffuseDef doubles from 46G to 96G, nevertheless, this number is close to the RSMI model as it requires gradient information during inference. Despite this increase, DiffuseDef is more efficient than ensembling-based methods like RanMask and SAFER, which require a full forward pass for all ensembles. With the same ensembling number of 10, both RanMask and SAFER require 459G FLOPS, which is 10x the number for the BERT baseline. In contrast, even with $t^\prime$ increased to 5, DiffuseDef can be run faster with 267G FLOPS, mitigating the efficiency problem from ensembling while maintaining the benefit of improved robustness.

Regarding training time, although training the diffusion layer requires more epochs, it costs similar time as training the classifier due to the small size of the diffusion layer. By contrast, adversarial training methods such as FreeLB++ and InfoBERT require much longer training time due to increased iterations of back-propagation per training batch.

\section{Conclusions}
We propose a novel adversarial defense method, DiffuseDef, which combines adversarial training, diffusion training, and ensembling to improve model robustness to adversarial attacks. DiffuseDef can build on any existing adversarial training method, training an additional diffusion layer to predict and remove randomly sampled noise at a given timestep. During inference, the diffusion layer is used to denoise the adversarial hidden states, which are ensembled to construct a robust text representation. Our experiments validate the effectiveness and efficiency of DiffuseDef, which significantly outperforms SOTA on common black-box and white-box adversarial attack methods, and maintains the performance on clean texts. Analysis shows that DiffuseDef makes it difficult to find vulnerable tokens to attack, and also reduces the difference between the hidden representations of clean and adversarial texts.

\section*{Limitations}
%In this work, we propose a new adversarial defense method, DiffuseDef, that incorporates existing adversarial training methods, diffusion training, and ensembling to improve robustness to adversarial attacks. 
%\paragraph{Scope} Our experiments focus on defending against three common black-box adversarial attack methods, while whether DiffuseDef improves model robustness against white-box attacks is unclear. \zlchange{White-box attacks have access to model parameters and can utilize gradient information to construct adversarial examples more efficiently than black-box attacks. Defending against white-box attacks is more challenging, and we consider this as a future direction of DiffuseDef.}

%\paragraph{Comparison with additional approaches} \zlchange{Due to the length limit, we do not compare against all current approaches. However we do compare with the SOTA methods with best adversarial robustness based on our preliminary experiments.}

%\paragraph{Efficiency} 
Despite the fact that DiffuseDef is more efficient than existing ensembling-based methods, it still requires more model parameters and inference FLOPS than non-ensembling-based models to achieve better robustness. Future directions of this work might involve efforts to reduce the size of the diffusion layer and the number of ensembles to make DiffuseDef more efficient. 

\section*{Ethical Considerations}
In this paper, we propose a new method DiffuseDef, which uses a diffusion layer as a denoiser to provide robust and efficient text representation. We demonstrate that the proposed method could significantly improve the robustness of NLP systems to adversarial attacks. However, DiffuseDef cannot defend against all adversarial attacks without limitations (e.g., number of perturbed words, semantic similarity between original and adversarial examples). Potential risks might include the creation of new adversarial attacks devised specifically for DiffuseDef.

\bibliography{anthology,custom}

% \clearpage
\appendix

\section{Data Preparation}
\label{sec:data_prep}
\begin{table}[htb]
    \centering
    \small
    \begin{tabular}{l|c|c|c|c}
    \toprule
    Dataset & Train & Valid & Test & Avg Len \\
    \midrule
    AGNews & 108K & 12K & 7K & 51.3 \\
    IMDB & 40K & 5K & 5K & 311.9 \\
    QNLI & 94K & 10K & 5K & 47.2 \\
    \bottomrule
    \end{tabular}
    \caption{Dataset statistics. The average text length is counted with BertTokenizer.}
    \label{tab:dataset_stats}
\end{table}
Table \ref{tab:dataset_stats} presents the number of examples in train/valid/test splits and the average token length for the three datasets used in the experiments. For QNLI and AGNews datasets, we randomly split the training set into our train/valid splits, with a ratio of 0.9/0.1, and use their test sets as our test split. For IMDB dataset, we randomly split the dataset into train/valid/test splits with a ratio of 0.8/0.1/0.1. All train/valid/test splitting is performed using a random seed of 42.

\section{Evaluation Constraints}
\label{sec:eval_param}
\begin{table}[htb]
    \centering
    \small
    \begin{tabular}{l|c|c|c}
    \toprule
    Dataset & $\mathbf{\varepsilon_{min}}$ & $\mathbf{K_{max}}$ & $\mathbf{\rho_{max}}$ \\
    \midrule
    AGNews & 0.84 & 50 & 0.3 \\
    IMDB & 0.84 & 50 & 0.1\\
    QNLI & 0.84 & 50 & 0.2 \\
    \bottomrule
    \end{tabular}
    \caption{Evaluation parameters for each dataset.}
    \label{tab:eval_params}
\end{table}
When evaluating with adversarial attack, we follow the parameter settings for TextAttack as suggested in \citep{li-etal-2021-searching}. The minimum semantic similarity $\mathbf{\varepsilon_{min}}$ between the clean text and adversarial text is set to 0.84, with the score computed using Universal Sentence Encoder \citep{cer-etal-2018-universal}. The maximum number of candidate substitutions $\mathbf{K_{max}}$ from the attacker is 50, thus the maximum number of queries $\mathbf{Q_{max}} = \mathbf{K_{max}} \times \mathbf{L}$ where $\mathbf{L}$ is the number of tokens. Finally, the maximum percentage of changed tokens $\mathbf{\rho_{max}}$ is set to 0.3/0.1/0.2 for AGNews, IMDB, and QNLI datasets respectively.

\section{Training}
\label{sec:training_param}
\begin{table}[htb]
    \centering
    \small
    \begin{tabular}{l|c c c}
    \toprule
     & AGNews & IMDB & QNLI \\
    \midrule
    Epochs & 100 & 100 & 100 \\
    Batch size & 64 & 64 & 64 \\
    Sequence len & 128 & 256 & 256 \\
    Dropout & 0.1 & 0.1 & 0.1 \\
    Optimizer & AdamW & AdamW & AdamW \\
    Lr & 2e-5 &  2e-5 & 2e-5 \\
    $t$ & 30 & 10 & 30 \\
    $t^\prime$ & 5 & 5 & 5 \\
    $k$ & 10 & 10 & 10 \\    
    \bottomrule
    \end{tabular}
    \caption{Hyperparameters for training DiffuseDef.}
    \label{tab:train_params}
\end{table}
The details on hyperparameters of diffusion training can be found in Table \ref{tab:train_params}. All models are trained on a single RTX A6000 GPU. The diffusion training of 100 epochs takes 6/4/3 hours on AGNews, IMDB, and QNLI datasets respectively.

\section{License for Scientific Artifacts}
\begin{table}[htb]
    \centering
    \resizebox{\linewidth}{!}{
    \begin{tabular}{l|c}
    \toprule
    Artifact & License \\
    \midrule
    AGNews \citep{NIPS2015_250cf8b5} & Custom (non-commercial) \\
    IMDB \citep{imdb} & - \\
    QNLI \citep{wang-etal-2018-glue} & CC BY-SA 4.0 \\
    \midrule
    transformers \citep{wolf-etal-2020-transformers} & Apache License 2.0 \\
    TextAttack \citep{morris-etal-2020-textattack} & MIT License \\
    \midrule
    BERT \citep{devlin-etal-2019-bert} & Apache License 2.0 \\
    RoBERTa \citep{liu2019roberta} & MIT License \\
    \bottomrule
    \end{tabular}
    }
    \caption{Licenses of scientific artifacts used in this paper.}
    \label{tab:licenses}
\end{table}
Table \ref{tab:licenses} lists the scientific artifacts, including data, codes, and models used in this paper. The use of these artifacts in this paper is consistent with their intended use, i.e., for scientific research only. The data used in the experiment is in English and does not contain personally identifying information or offensive content.

\section{Robustness with RoBERTa}
\label{sec:results_roberta}
In Table \ref{tab:main_results_roberta}, we report the black-box attack results on RoBERTa-based models. A similar conclusion holds for BERT-based models, and DiffuseDef significantly outperforms other defense methods on all 3 datasets.
\begin{table*}[htb]
    \centering
    \small
    \resizebox{.93\linewidth}{!}{
    \begin{tabular}{c|c|l|c|c c c|c c c}
    \toprule
    % Table top rows
    \multirow[m]{2}{*}{ \textbf{Dataset} } & \multirow[m]{2}{*}{ \textbf{PLM} } & \multirow[m]{2}{*}{ \textbf{Method} } & \multirow[m]{2}{*}{\textbf{Clean\%}} & \multicolumn{3}{c|}{\textbf{AUA\%}} & \multicolumn{3}{c}{\textbf{\#Query}} \\ 
    & & & & TF & TB & BA & TF & TB & BA \\
    \midrule 
    \multirow{6}{*}{ AGNews } & \multirow{6}{*}{ RoBERTa-base } & Fine-Tuned & 94.9 & 34.1 & 36.9 & 43.6 & 372 & 396 & 410 \\
     & & InfoBERT & 95.5 & 40.2 & 45.2 & 48.6 & 392 & 421 & 430 \\
     & & FreeLB++ & 95.4 & 57.5 & 62.9 & 55.9 & 444 & 467 & 447 \\
     & & RSMI & 93.1 & 64.2 & 66.4 & 67.4 & 774 & 861 & 808 \\
     & & DiffuseDef-FreeLB++ (Ours) & 95.3 & \textbf{85.6} & \textbf{87.6} & \textbf{85.3} & 880 & \textbf{976} & 906 \\
     & & DiffuseDef-RSMI (Ours) & 92.9 & 82.9 & 83.5 & 82.2 & \textbf{905} & 925 & \textbf{1047} \\
    \midrule
    \multirow{6}{*}{ IMDB } & \multirow{6}{*}{ RoBERTa-base } & Fine-Tuned & 94.6 & 21.3 & 17.9 & 13.6 & 587 & 671 & 493 \\
     & & InfoBERT & 94.8 & 30.9 & 27.9 & 21.8 & 681 & 760 & 549 \\
     & & FreeLB++ & 95.3 & 46.0 & 42.1 & 33.9 & 829 & 974 & 637 \\
     & & RSMI & 92.7 &  77.9 & 74.3 & 70.6 & 3443 & 4342 & 2619 \\
     & & DiffuseDef-FreeLB++ (Ours) & 95.0 & \textbf{86.2} & \textbf{85.9} & \textbf{86.8} & 3573 & 4663 & 2941 \\
     & & DiffuseDef-RSMI (Ours) & 92.4 & 84.7 & 84.1 & 84.3 & \textbf{3673} & \textbf{4782} & \textbf{3007} \\
    \midrule
    \multirow{6}{*}{ QNLI } & \multirow{6}{*}{ RoBERTa-base } & Fine-Tuned & 92.8 & 26.6 & 22.5 & 20.3 & 204 & 219 & 188 \\
     & & InfoBERT & 92.5 & 29.1 & 25.8 & 20.3 & 205 & 223 & 189 \\
     & & FreeLB++ & 92.8 & 33.9 & 27.7 & 20 & 227 & 244 & 200 \\
     & & RSMI & 89.3 & 38.8 & 33.2 & 30.0 & 340 & 375 & 343 \\
     & & DiffuseDef-FreeLB++ (Ours) & 92.7 & 64.6 & 64.3 & 61.5 & 473 & 579 & 524 \\
     & & DiffuseDef-RSMI (Ours) & 88.8 & 57.7 & 55.7 & 53.9 & 469 & 578 & 518 \\
    \bottomrule
    \end{tabular}
    }
    \caption{Main adversarial robustness results on classification tasks with RoBERTa. \textit{Clean}: accuracy on clean test set. \textit{TF}: TextFooler. \textit{TB}: TextBugger. \textit{BA}: BertAttack.}
    \label{tab:main_results_roberta}
\end{table*}

\section{Robustness w.r.t Token Length}
\begin{figure}[htb]
    \centering
    \includegraphics[width=\linewidth]{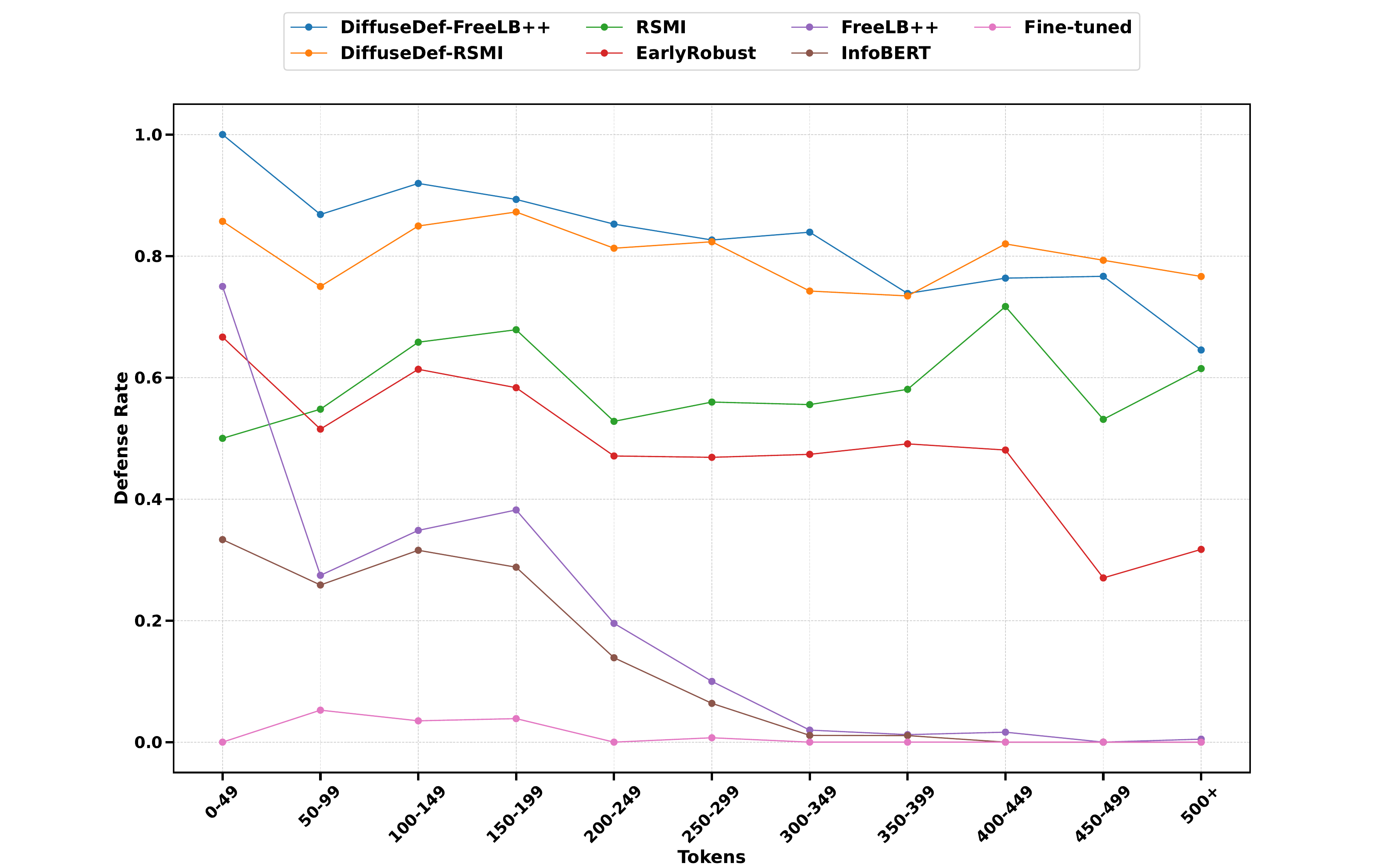}
    \caption{Defense rate (against TextFooler) w.r.t token length for different models on IMDB dataset.}
    \label{fig:defense_rate_length}
\end{figure}
Figure \ref{fig:defense_rate_length} provides a comparison of the defense rate for different models by token length on the IMDB dataset. The defense rate is calculated as the percentage of test examples in which TextFooler fails to construct a successful attack. All models except RSMI show a consistent trend that the defense rate declines as the texts lengthen. This trend can be attributed to the nature of adversarial attacks, as longer texts allow for the generation of more adversarial examples. Specifically, adversarial training defense methods like InfoBERT and FreeLB++ show poor performance on longer texts (more than 300 tokens), with the defense rate reduced to nearly 0. This drastic decline indicates that, given an adequate number of queries, the attacker is guaranteed to find a successful attack to fool these models. Similarly, EarlyRobust exhibits a performance drop on long texts as it is based on FreeLB training. RSMI, however, performs worse on short texts, but its defense rate increases as the text length grows. Compared to all SOTA defense approaches, the two DiffuseDef variants show a more steadily declining trend and maintain a higher defense rate across all token lengths, i.e., DiffuseDef is more robust to input text length.

\section{Example of noising and denoising in DiffuseDef}
Adding and removing noise to hidden states are essential features in DiffuseDef, which contribute to the improved adversarial robustness. To study how adding or removing noise can affect the semantic meaning of the text, we feed the hidden states to the pretrained BERT model with a masked language modeling (MLM) head to generate the text output. 

In Table \ref{tab:nosie_example}, we present the MLM outputs from hidden states added with different steps of noise and the MLM outputs from noise hidden states denoised with the same number of steps.  In the example shown, with more noise added some semantic information can be lost and replaced by symbols or function words like ``.'' or ``the''. In contrast, denoising for the same number of steps helps alleviate such information loss. For example, the word ``IBM'' can be recovered from the noise.

However, in practice, it is not possible to assume the number of denoising steps; therefore in Table \ref{tab:denosie_example} we show the MLM outputs of denoised hidden states directly from clean and adversarial texts. On clean text, we observe that a higher number of denoising steps can result in more abstraction of the texts. For example, more words are replaced with ``the'' in the MLM outputs as $t^\prime$ grows. However, words related to the topic (e.g., ``Manchester United'', ``Liverpool'') are kept during the denoising process, thus the model can predict correctly. Similarly, the trend of abstraction can also be found on adversarial text while we observe that the denoising can help remove the adversarial noise/perturbation and recover the word ``united'' from ``nation'', thus resulting in its correct prediction on the adversarial text.
\begin{table*}[htb]
    \centering
    \small
    \resizebox{\linewidth}{!}{
    \begin{tabular}{c p{0.47\textwidth} p{0.47\textwidth}}
    \toprule
    % Table top rows
    $\mathbf{t^\prime}$ & \textbf{MLM Output (add noise)} & \textbf{MLM Output (add noise then denoise)} \\ 
    \midrule 
    0 & IBM Chips May Someday Heal Themselves New technology applies electrical fuses to help identify and repair faults. & - \\ \midrule
    5 & the ibm chips may someday heal themselves new technology \hlchange{introduces} electrical fuses to help identify and repair faults. & the ibm chips may someday heal themselves new technology \hlchange{introduces} electrical fuses to help identify and repair faults. \\ \midrule
    6 & \hlchange{)} ibm chips may someday heal themselves new technology \hlchange{introduces} electrical fuses to help identify and repair faults. & the ibm chips may someday heal themselves new technology \hlchange{introduces} electrical fuses to help identify and repair faults. \\ \midrule
    7 & \hlchange{the.} chips may someday heal themselves new technology \hlchange{introduces} electrical fuses to help identify and repair faults. & \hlchange{the.} chips may someday heal themselves new technology \hlchange{uses} electrical fuses to help identify and repair faults. \\ \midrule
    8 & \hlchange{)..} may someday heal themselves new technology \hlchange{introduces} electrical fuses to help identify and repair faults. & \hlcorrect{the ibm.} may someday heal themselves new technology \hlchange{uses} electrical fuses to help identify and repair faults. \\ \midrule
    9 & the ibm chips may someday heal themselves new technology \hlchange{uses} electrical fuses to help identify and repair faults. & the ibm chips may someday heal themselves new technology \hlchange{introduces} electrical fuses to help identify and repair faults. \\ \midrule
    10 & \hlchange{the.} chips may someday heal themselves new technology \hlchange{extends} electrical fuses to help identify and repair faults. & \hlcorrect{the ibm.} may someday heal themselves new technology \hlchange{develops} electrical fuses to help identify and repair faults. \\
    \bottomrule
    \end{tabular}
    }
    \caption{MLM outputs from hidden states with noise added and hidden states with first noise added but then denoised. We only report $t^\prime$ above 5 as the MLM outputs with smaller $t^\prime$ are identical to the clean text.}
    \label{tab:nosie_example}
\end{table*}

\begin{table*}[htb]
    \centering
    \small
    \begin{tabular}{c p{0.34\textwidth} p{0.34\textwidth} c c}
    \toprule
    % Table top rows
    $\mathbf{t^\prime}$ & \textbf{Clean Text / MLM Output} & \textbf{Adv Text / MLM Output}  & \textbf{Pred clean} & \textbf{Pred adv} \\ 
    \midrule 
    0 & United Apology over Website Abuse Manchester United have been forced to issue an embarrassing apology to Liverpool for an ill-advised attack on the Anfield outfit on its own website. & United Apology over Website Abuse Manchester \hladv{Nations} have been forced to issue an embarrassing apology to Liverpool for an ill-advised attack on the Anfield outfit on its own website. & Sports & \textcolor{red}{World}\\ \midrule
    1 & \hlchange{football.} apology over website abuse manchester united have been \hlchange{-} to issue an embarrassing apology to liverpool for an \hlchange{the -} advised attack on the anfield outfit on its own website. & \hlchange{the.} apology over website abuse manchester \hladv{nations} have been \hlchange{the} to issue an embarrassing apology to liverpool for an \hlchange{the -} advised attack on the anfield outfit on its own website. & Sports & \textcolor{red}{World} \\ \midrule
    2 & \hlchange{the.} apology over website abuse manchester united have been \hlchange{-} to issue an embarrassing apology to liverpool for an \hlchange{the -} advised attack on the anfield outfit on its own website. & \hlchange{the.} apology over website abuse manchester \hladv{nations} have been \hlchange{the} to issue an embarrassing apology to liverpool for an \hlchange{the -} advised attack on the anfield outfit on its own website. & Sports & \textcolor{red}{World} \\ \midrule
    3 & \hlchange{the.} apology over website abuse manchester united have been \hlchange{the} to issue an embarrassing apology to liverpool for an \hlchange{the -} advised attack on the anfield outfit on its own website. & \hlchange{the.} apology over website abuse manchester \hlchange{s} have been \hlchange{the} to issue an embarrassing apology to liverpool for an \hlchange{the -} advised attack on the anfield outfit on its own website. & Sports & Sports \\ \midrule
    4 & \hlchange{the.} apology over website abuse manchester united have \hlchange{the -} to issue an embarrassing apology to liverpool for an \hlchange{the -} advised attack on the anfield outfit on its own website. & \hlchange{the.} apology over website abuse manchester \hlchange{s} have been \hlchange{the} to issue an embarrassing apology to liverpool for an \hlchange{the -} advised attack on the anfield outfit on its own website. & Sports & Sports \\ \midrule
    5 & \hlchange{the.} apology over website abuse manchester united have \hlchange{the the} to issue an a apology to liverpool for an \hlchange{the -} advised attack on the anfield outfit on its own website. & \hlchange{the.} apology over website abuse manchester \hlcorrect{united} have been \hlchange{the} to issue an the apology to liverpool for \hlchange{an'-} advised attack on the anfield outfit on its own website. & Sports & Sports \\
    \bottomrule
    \end{tabular}
    \caption{MLM outputs and FreeLB++ model predictions from ensembling diffused hidden states at different denoising steps.}
    \label{tab:denosie_example}
\end{table*}

\section{Confusion Matrix under Attack}
\begin{figure*}[t]
\centering
  \includegraphics[width=\textwidth]{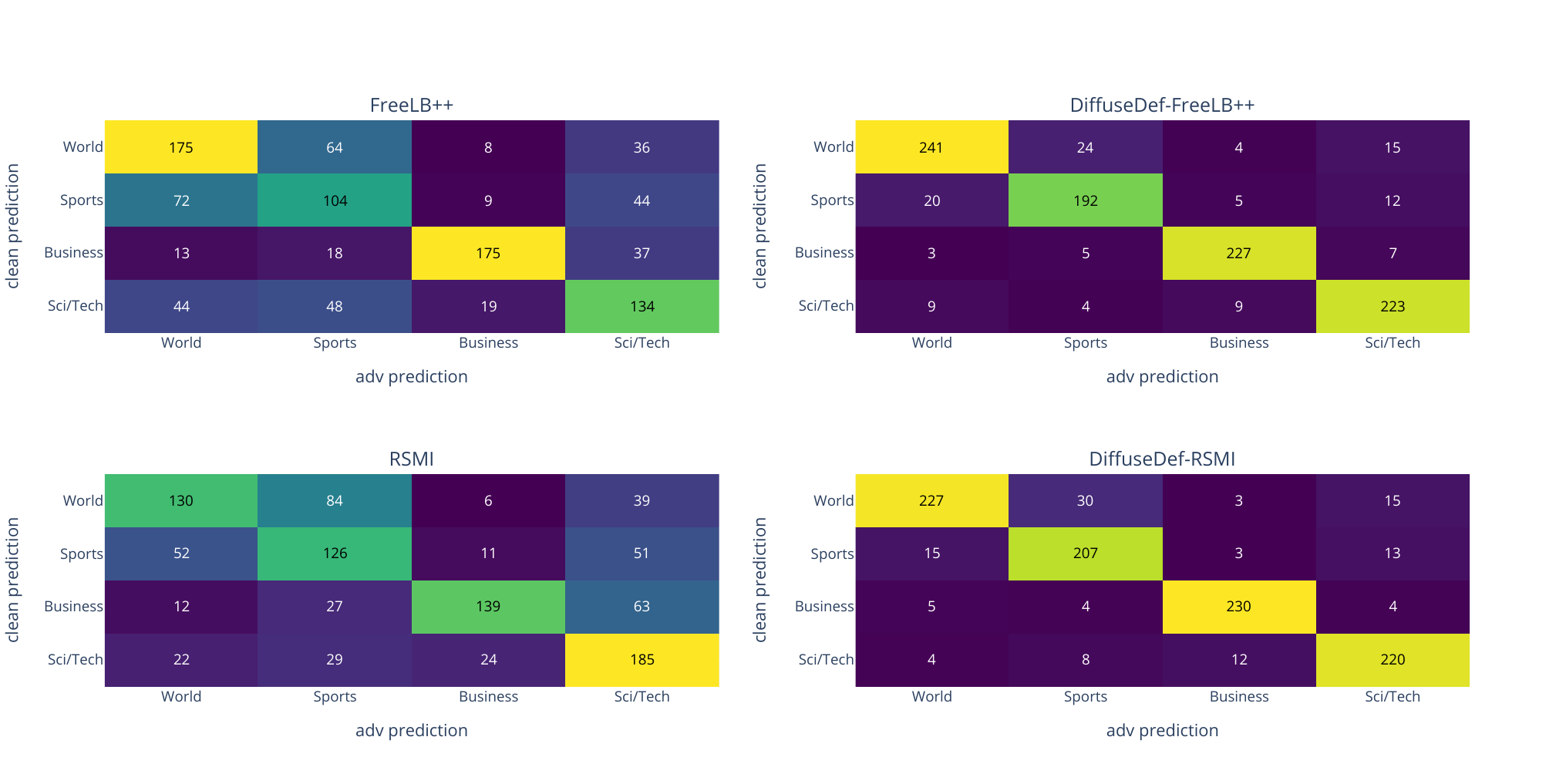}
  \caption{Confusion matrix of models under attack on AGNews test set.}
  \label{fig:cm_ag}
\end{figure*}

\begin{figure*}[b]
\centering
  \includegraphics[width=\textwidth]{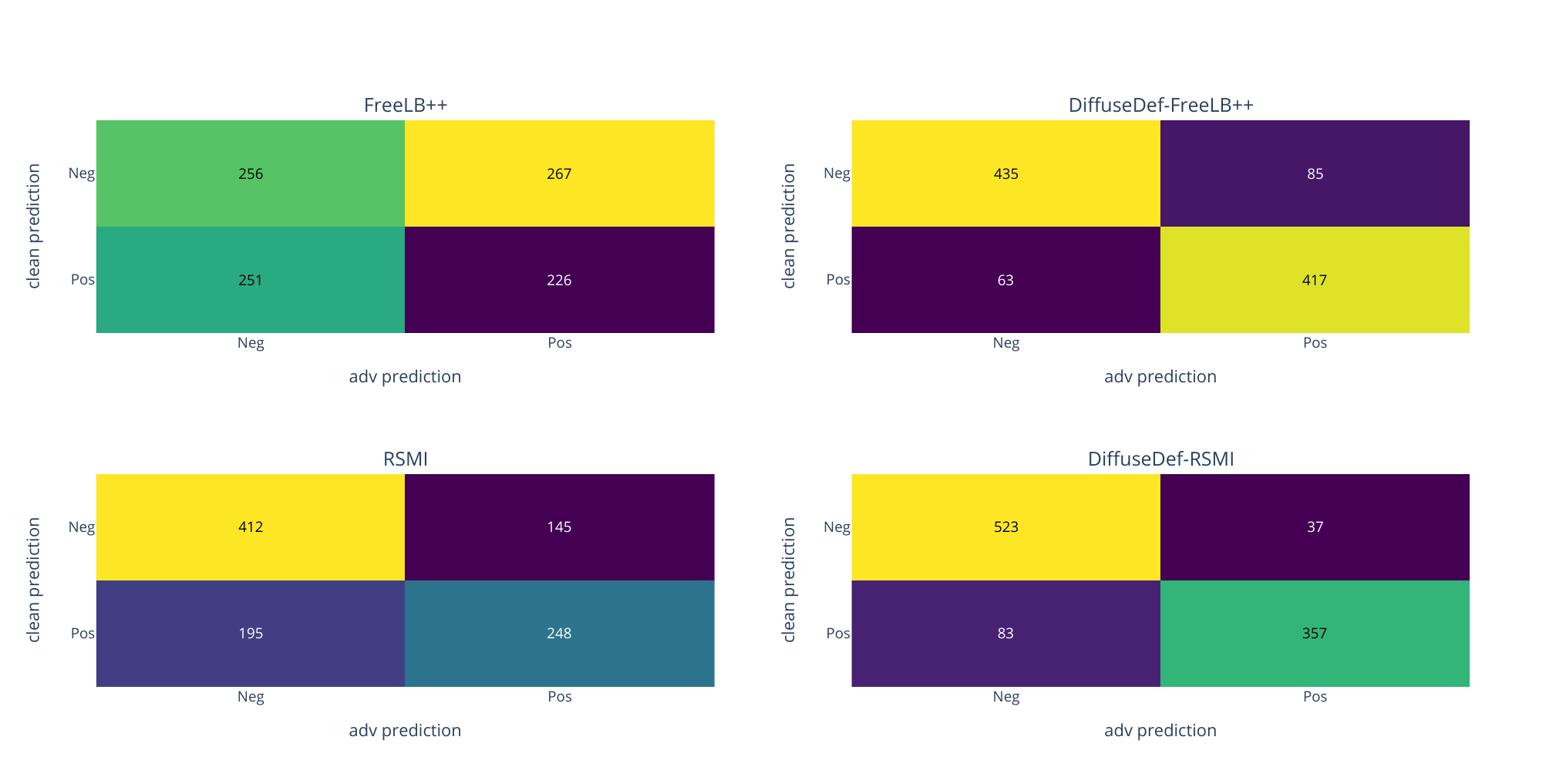}
  \caption{Confusion matrix of models under attack on IMDB test set.}
  \label{fig:cm_imdb}
\end{figure*}

Figure \ref{fig:cm_ag} and \ref{fig:cm_imdb} present the confusion matrices of models' predictions on clean text and on adversarial texts (successful attack example) on AGNews and IMDB test sets respectively. 

\end{document}